\pdfoutput=1
\documentclass[10pt]{article}

\usepackage[preprint]{tmlr}

\usepackage[utf8]{inputenc}
\usepackage[T1]{fontenc}
\usepackage{url}
\usepackage{booktabs}
\usepackage{longtable}
\usepackage{amsfonts}
\usepackage{nicefrac}
\usepackage{microtype}
\PassOptionsToPackage{table}{xcolor}
\usepackage{xcolor}
\definecolor{linkblue}{RGB}{0,0,120}
\usepackage{graphicx}
\usepackage{array}
\usepackage{tabularx}
\usepackage{enumitem}
\usepackage{tikz}
\usetikzlibrary{arrows.meta,positioning,shapes.geometric,fit,backgrounds}
\usepackage{placeins}
\usepackage{float}
\usepackage{flafter}
\usepackage{hyperref}
\hypersetup{
  colorlinks=true,
  citecolor=linkblue,
  linkcolor=linkblue,
  urlcolor=linkblue,
  anchorcolor=linkblue,
  breaklinks=true
}

\title{Designing Benchmarks for Knowledge Work}
\author{\name Yining Hua\thanks{Corresponding author: Yining Hua (\href{mailto:yininghua@g.harvard.edu}{yininghua@g.harvard.edu}).} \\
      \addr Harvard University, Cambridge, MA, USA
      \AND
      \name Hongbin Na \\
      \addr University of Technology Sydney, Sydney, Australia
      \AND
      \name Cyrus Ayubcha \\
      \addr Harvard University, Cambridge, MA, USA
      \AND
      \name Levi Lian \\
      \addr Raycaster \\
      \addr Stanford University, Stanford, CA, USA}

\def\month{MM}
\def\year{YYYY}
\def\openreview{\url{https://openreview.net/forum?id=XXXX}}

\begin{document}

\maketitle
\begin{abstract}
Agent benchmarks have attracted substantial attention; yet, benchmarks for AI agents performing knowledge work are still mostly described only by their tasks, environments, and metrics. Starting from the definition of knowledge work, this paper derives four other design choices that often remain implicit. Three research threads on how knowledge work is organized each yield one choice: studies of professions, situated action, and boundary objects and coordination, while validity theory provides the fourth. Following this reasoning, we propose a benchmark representation centered on the work being performed. Based on this representation, we propose reporting practices for four largely missing fields: represented activity, tested setting, required work product, and evaluated result. Benchmarks designed in disparate ways thereby share a common description and can be compared side by side. To support the reporting of work activities, we cluster O{*}NET task statements into 18 shared work activities through unsupervised learning paired with human review. We then use this representation to analyze cases from three popular knowledge work benchmarks: GDPval, OfficeQA Pro, and APEX-SWE. Across all three benchmarks, the scope covered by the evaluated result is narrower than that of the required work product. We audited the tests of all 25 public APEX-SWE Integration tasks, independently coded by two authors. We found from the test code that in 13 of these tasks, a deliverable can alter established content from the source data and still pass all tests. We synthesize these four fields into a one-page record, accompanied by a worked example and a fillable template.
\end{abstract}

\section{Introduction}
\label{sec:intro}

There is a growing narrative regarding the capability of agents to perform knowledge work. For example, OpenAI reported that 47.6\% of the outputs from the top-performing model at that time, i.e., Claude Opus 4.1, were rated as better than or equal to the work of industry experts on GDPval \citep{openai2025gdpval}. Artificial Analysis now also lists knowledge-work benchmarks separately alongside reasoning and coding evaluations, such as GDPval-AA and APEX-Agents-AA \citep{artificialanalysis2026}. However, we observe that the design and reporting methodologies of these benchmarks are insufficient to reflect actual work competence. Flaws in task setup or reward design can cause these agentic benchmarks to overestimate or underestimate performance by up to 100\% in relative terms \citep{zhu2025establishing}; furthermore, work-oriented benchmarks are still primarily described using only tasks, environments, and metrics.

We posit that the cause of this insufficiency of reporting is a lack of evaluation guidelines made specifically for knowledge work; instead, researchers continue to rely on guidelines designed for tasks that previous LLMs could perform. Conventional LLM benchmarks primarily evaluate capabilities, such as answering questions based on a given text \citep{rajpurkar2016squad} or retrieving relevant documents \citep{thakur2021beir}. For a capability, matching a short answer against a reference answer serves as valid evidence. Work is different. Knowledge work is labor where knowledge serves as both the raw material and the output \citep{drucker1999productivity,davenport2005thinking}, accounting for a substantial share of the labor market \citep{porat1977information}. The same output can hold entirely different meanings across different roles and workflows \citep{abbott1988system}. Furthermore, an output is only considered complete when others can inspect or utilize it in subsequent steps \citep{carlile2004transferring}. All these considerations should inform the design and reporting of knowledge work benchmarks. 

Such labor is performed by a specific role, and its work product must be handed off for the next person to build upon (Section~\ref{sec:knowledge-work}). Agents carrying out this type of work use tools and execute multi-step workflows \citep{wang2024survey}. Consequently, a score on a work task may be based either on a short answer \citep{browsecomp2024} or on modifications made to a running system \citep{kottamasu2026apex}, and these two scores support very different claims about work. Therefore, a work-oriented benchmark needs to make two things explicit: which work it represents, and which part of that work its score evaluates. Current practice leaves both implicit.

In this paper, we start from investigating how knowledge work is defined and organized, to systematically derive four minimal essential design choices that a work-oriented benchmark should clarify. To the best of our knowledge, this is the first study to derive benchmark design fields from this body of research: three research threads on how knowledge work is organized each yield one field, while validity theory provides the fourth \citep{abbott1988system,suchman1987plans,star1989institutional,messick1989validity}. Following this reasoning, we propose a work-centric benchmark representation that records these four choices as four independent fields: \emph{represented activity}, \emph{tested setting}, \emph{required work product}, and \emph{evaluated result}.

\begin{figure}[!t]
\centering
\includegraphics[width=\textwidth]{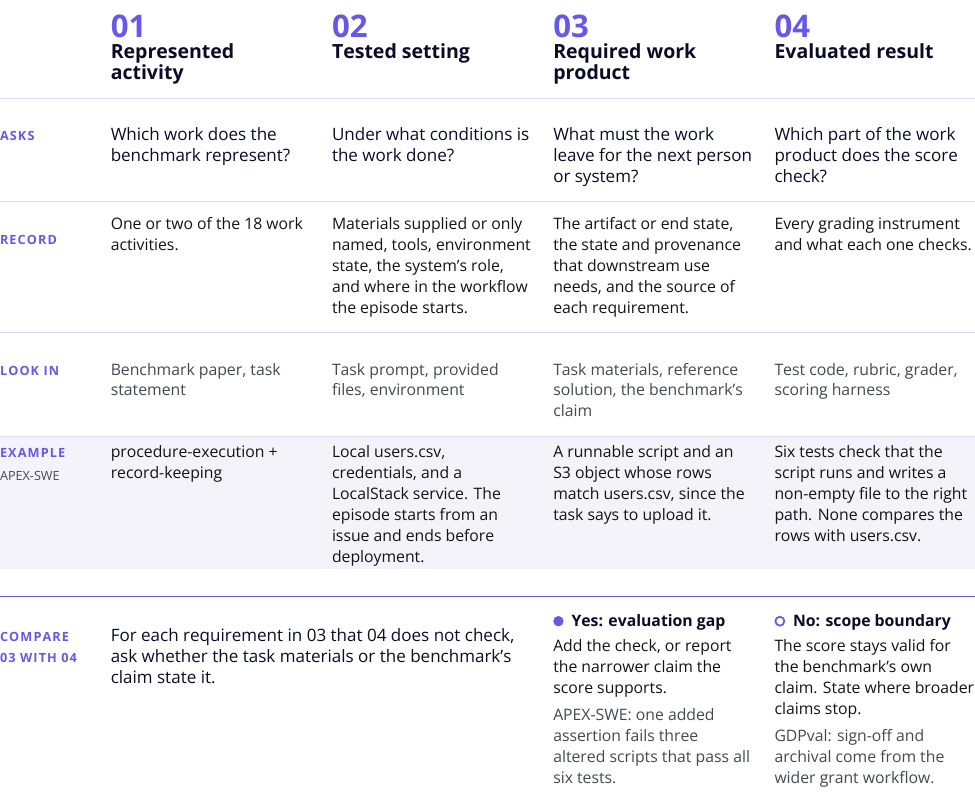}
\caption{The four-field record on one page. Each column gives the question a field answers, what to record, where to look, and an example from the APEX-SWE task in Section~\ref{sec:whybench}. The 18 work activities are listed in Table~\ref{tab:ops}. The bottom row applies the rule in Section~\ref{sec:score-work-product} to requirements that the evaluated result does not check. Table~\ref{tab:case-comparison} fills the record for three benchmarks, and a fillable template is in the supplementary material.}
\label{fig:card}
\end{figure}

Prior analyses have treated benchmark tasks as a single monolithic unit, whether tracking the chain from phenomena to conclusions \citep{bean2025measuring} or mapping benchmarks directly to occupations \citep{wang2026realwork}. Our representation separates the output required by the task from the part actually evaluated by the benchmark. In educational measurement, Evidence-Centered Design (ECD) similarly distinguishes between tasks, work products, and scoring \citep{mislevy2003structure}. However, in ECD, work products are defined in terms of how much evidence they provide about the test-taker. In contrast, we define the required work products of a task based on the actual needs of the downstream person or system receiving the work. We also propose a decision rule that distinguishes evaluation gaps from scope boundaries based on the origin of each requirement (Section~\ref{sec:score-work-product}). To support the reporting of work activities, we also cluster O{*}NET task descriptions into 18 cross-occupational work activities using unsupervised learning combined with human verification. This representation serves a role for work-oriented benchmarks analogous to what datasheets \citep{gebru2021datasheets} and model cards \citep{mitchell2019modelcards} do for datasets and models. Figure~\ref{fig:card} gives the four fields on one page, with a worked example. We apply this representation to case studies of GDPval, OfficeQA Pro, and APEX-SWE, and audit all 25 public Integration tasks in APEX-SWE. In 13 of these tasks, the test code shows that an output with wrong data (e.g., made-up records) can still pass all the released tests. 

\begin{table}[!ht]
\caption{Definitions used in this paper.}
\label{tab:terms}
\centering
\footnotesize
\setlength{\tabcolsep}{2pt}
\setlength{\extrarowheight}{1pt}
\renewcommand{\arraystretch}{1.1}
\begin{tabularx}{\linewidth}{@{}>{\raggedright\arraybackslash}p{0.15\linewidth}>{\raggedright\arraybackslash}X@{}}
\toprule
\textbf{Term} & \textbf{Definition} \\
\midrule
Knowledge work &
A form of labor in which knowledge is both the source material and the output \citep{drucker1999productivity,davenport2005thinking}. \\
Work activity &
A recurrent unit of work that can appear across occupations \citep{onet2026overview,abbott1988system}. \\
Benchmark task &
The benchmark-defined task assigned to a system, including its inputs, available actions, environment, and expected submission \citep{bean2025measuring,liang2023helm}. \\
Tested setting &
The benchmark-defined conditions under which the work activity is evaluated, including materials, tools, role, and workflow state \citep{suchman1987plans,hutchins1995cognition}. \\
Work product &
The object left for review, filing, execution, or downstream continuation, including any state or provenance needed for that use \citep{star1989institutional,carlile2004transferring}. \\
Evaluated result &
The part of the submitted output, artifact, or resulting state that the benchmark checks through its metric, tests, or rubric \citep{bean2025measuring,kane2013validating}. \\
\bottomrule
\end{tabularx}
\end{table}

\section{What Knowledge Work Requires}
\label{sec:knowledge-work}

\subsection{Labor: What Kind of Work a Task Represents}
\label{sec:labor}

Knowledge work is first and foremost a form of labor, and labor is divided across occupations \citep{freidson2001professionalism}. Abbott argues that this division of labor constitutes a division of jurisdiction over different types of problems; the same class of problem can emerge across multiple occupations and can shift from one occupation to another. Moreover, the nature of a task is partially defined by the occupation that handles it: the same matter can be framed and treated as different types of problems, and how it is categorized determines what it should look like and by what standards it is judged \citep{abbott1988system}. Consequently, tasks sharing identical forms can belong to entirely different jobs. For example, compiling a numerical table from a set of reports could serve as the final step of an analysis or as a verification check during an audit. Conversely, an occupation is an amalgamation of many tasks \citep{autor2003skill,acemoglu2011skills} that span multiple lines of work with distinct criteria of competence.

Therefore, specifying only tasks or occupations fails to clarify what kind of work a benchmark actually represents. From this, we arrive at the first field, \textbf{represented activity}, which documents the type of work a benchmark represents. This unit of analysis is larger than an individual task yet smaller than an occupation, recurring across different professions; we term it a \textbf{work activity} \citep{peterson2001onet}. An activity is demarcated by the transformations it performs and the outputs it produces. For example, \emph{analysis} turns evidence into substantiated conclusions, whereas \emph{record-keeping} establishes, updates, or retrieves official records. From the task statements in O{*}NET, we synthesize 18 categories of activities (Section~\ref{sec:vocab}).

\subsection{Input: What Work Relies Upon}
\label{sec:input}

Knowledge work relies on knowledge as its input. Suchman's research on situated action demonstrates that plans merely serve as resources for action rather than determining its course; people organize their actions by relying on the materials and context immediately at hand \citep{suchman1987plans}. Hutchins's study of ship navigation teams further reveals that cognition is distributed across people, tools, and records: what each member knows and can do depends on their position within the team and the instruments before them \citep{hutchins1995cognition}. Consequently, a substantial portion of the knowledge mobilized in knowledge work resides in documents, systems, and colleagues, varying across different workplaces and even across different stages of the same workflow \citep{hollan2000distributed,orlikowski2002knowing}. For example, drafting a grant review evaluation after receiving a curated dossier versus retrieving documents independently from an entire archive constitutes two fundamentally distinct kinds of work, as the evidence in the former has already been filtered and selected; likewise, an evaluation written directly by the decision-maker involves distinct work from one drafted by an assistant for that decision-maker.

In other words, the same activity becomes a different job once the materials, tools, and roles change \citep{lave1988cognition,engestrom2001expansive}. This gives rise to the second field, \textbf{tested setting}, which records the materials and tools the benchmark supplies to the system, the role it assigns to the system, and the step in the workflow from which the system begins.

\subsection{Output: What Work Must Leave Behind}
\label{sec:output}

The output of knowledge work is likewise knowledge, typically intended for subsequent use by another person or system. Carlile differentiates three boundaries across which knowledge must cross: when parties share a common syntax and background, transferring information is sufficient; when parties interpret the same content differently, knowledge must be translated; and when parties have conflicting interests such that one party's new knowledge imposes costs on another, knowledge must be transformed \citep{carlile2002pragmatic,carlile2004transferring}. Star and Griesemer's concept of the boundary object illustrates the nature of outputs that circulate effectively across distinct groups: they must be adaptable to local constraints within each party's context while maintaining a common identity across them \citep{star1989institutional}. Coordination theory conceptualizes the relationship where ``one activity produces something that is used by another activity'' as a fundamental dependency between activities; managing this dependency involves ensuring that the output can be directly utilized by the downstream activity \citep{malone1994coordination,thompson1967organizations}. For instance, an accurate figure provided without provenance cannot be verified by a reviewer, and a database update that fails to log which fields were altered cannot be audited downstream \citep{buneman2001why}.

Whether a piece of work is complete therefore depends on whether the next person can inspect and utilize what it leaves behind, rather than solely on whether its contents are correct \citep{okhuysen2009coordination}. This yields the third field, \textbf{required work product}, which specifies the artifact that the work must produce, including the state and provenance information necessary for downstream use.

\subsection{Score: What Was Evaluated}
\label{sec:score}

Benchmarks measure outputs with scores to facilitate comparisons across different agents. People also draw conclusions about work based on these scores, such as ``this agent is competent in managing federal grants.'' Whether such a conclusion holds is a question of validity \citep{aera2014standards}. Messick argues that validity is not a property of the test itself, but the degree to which empirical evidence and theoretical rationales support the interpretations drawn from test scores \citep{messick1989validity}. Kane structures the progression from an agent's performance to the ultimate conclusion into a chain of inferences, where each step requires evidence \citep{kane2013validating}. The first step in this chain is scoring: deriving a score from a work product. Yet scores can only reflect those properties actually checked by metrics, tests, or rubrics \citep{mislevy2003structure}.

In knowledge work, scoring often checks only a fraction of a work product. A work product typically comprises multiple parts, such as a script alongside the resulting system state after execution, or a figure accompanied by the records supporting it. Nevertheless, scoring may only check whether the script runs or whether the figure is correct. Consequently, scores provide no evidence regarding unchecked components, and conclusions drawn from scores cannot extend to them \citep{messick1995validity}. This reasoning yields our fourth field, \textbf{evaluated result}, which records the subset of the required work product that is actually inspected and separates it from the product itself.

\section{A Work-Centered Representation for Benchmark Design and Reporting}
\label{sec:guidelines}

This section explains how each field should be reported in turn, and introduces an O{*}NET-based vocabulary for the first field, represented activity. Figure~\ref{fig:card} summarizes the four fields.

\subsection{Represented Activity}
\label{sec:define-work}

The represented activity field should document the work that the benchmark intends to represent. For a work-oriented benchmark, the activity serves as the target construct; a benchmark task is an operationalization of this activity under a given tested setting, and a single episode may also encompass multiple activities \citep{adcock2001measurement,jacobs2021measurement}. The activity label specifies what work is being performed, while benchmark reporting further articulates how the task operationalizes it and which aspects fall outside the evaluation.

Currently, benchmark labels typically fall into one of three tiers, none of which clearly identifies the represented activity. Domain or occupational labels situate a benchmark within an applied area, but each label subsumes a diverse set of activities with differing standards of competence \citep{handel2016onet,onet2024tasks}. Consequently, a benchmark may sample only a small fraction of an occupation while carrying the name of the entire profession \citep{raji2021everything}. Component labels isolate narrower behaviors \citep{rajpurkar2016squad,nallapati2016cnndm,thakur2021beir}; retrieval, for instance, can take place within many larger activities. Evaluating performance on a component alone cannot determine which activity it represents, as that depends on how the component is deployed within the tested setting and the work product. Agent task labels bundle multiple actions into a single episode \citep{zhou2023webarena,xie2024osworld,wang2024officebench}, where search and computation may both occur before a final submission; an aggregate score leaves unclear which activities the work product actually requires and which actions arise merely as artifacts of the environment.

Work-oriented reporting maps an episode to one or more activities, separating the activities genuinely demanded by the work product from incidental actions. Activity labels are intentionally kept coarse-grained. Tasks within the same activity may differ in their surrounding workflows and required deliverables; these variations are captured in the tested setting and required work product fields rather than through further sub-labeling. Two analysis tasks---one concluding with a numerical answer and the other with a reviewable memo---share the same activity label, with their distinctions documented under work product. When the boundaries between activities are loose, an episode can carry two labels simultaneously, as demonstrated in our APEX-SWE case study in Section~\ref{sec:whybench}.

\subsection{Tested Setting}
\label{sec:tested-setting}

The tested setting field records the conditions under which the activity is evaluated, including (1) the materials available to the system and their initial state, (2) the tools available to the system and the persistent environmental state left behind after execution \citep{zhou2023webarena,xie2024osworld,drouin2024workarena}, (3) the role assigned to the system by the benchmark, and (4) the workflow stage at which the episode commences. Execution environment specifications typically report only which services and files are present; the tested setting further specifies which materials are authoritative, distinguishing materials directly supplied by the benchmark from sources that the task merely references and expects the system to retrieve independently (e.g., external statistics cited but not attached). Upstream choices made by the benchmark on behalf of the system also constitute part of the tested setting: providing a curated set of source materials pre-determines the evidence selection, while prescribing output formats pre-determines the artifact's presentation.

Role refers to the persona and responsibilities that the benchmark assigns to the system, typically articulated in the task prompt (e.g., instructing the system to act as a federal grant manager). It is documented separately from the represented activity because the same activity can be performed either by a primary decision-maker or by an assistant drafting on their behalf, each imposing distinct requirements on the work product.

The tested setting should also clarify whether this specific setting constitutes the intended target of the benchmark's claims or represents a simplification of a broader class of scenarios \citep{kane2013validating}. SWE-bench is constructed from real GitHub issues \citep{jimenez2024swebench} and evaluates issue resolution given a codebase snapshot and corresponding tests, leaving code review and deployment outside the episode. Benchmarks may intentionally evaluate only a single stage of a workflow or restrict tooling; explicitly reporting these choices establishes the valid scope of claims supported by the resulting scores.

\subsection{Required Work Product}
\label{sec:score-work-product}

The required work product specifies the artifacts or system state that the activity is expected to yield in the tested setting (e.g., a revised document or a database update), including the state and provenance information required for downstream use. The activity label also informs what the product ought to preserve; Appendix~\ref{app:opslens} catalogs these elements for each activity class. When designing a benchmark, the product is dictated by the intended claims regarding the work and the downstream workflow that receives it \citep{mislevy2003structure}. For example, a benchmark evaluating source-grounded document revision should specify the revised draft delivered to reviewers and identify which sections must remain traceable to sources. When analyzing existing benchmarks, specifications are drawn primarily from the benchmark's claimed conclusions and task instructions.

Requirements on the work product stem from three sources. The first is task-specified requirements, explicitly stated in the task materials, including reference solutions; in the APEX-SWE task analyzed in Section~\ref{sec:whybench}, the prompt instruction ``upload local \texttt{users.csv}'' makes row-by-row parity between the file and the uploaded entities a required work product specification. The second is claim-specified requirements, derived from properties directly invoked by the benchmark's claims; OfficeQA Pro claims to evaluate end-to-end grounded reasoning, making correct use of source data a required work product property even if the task prompt asks only for a final numeric value. The third is analytic inference, derived from workflows implied by the benchmark's broader claims but left unspecified in its design. GDPval claims to evaluate realistic, economically valuable knowledge work; one of its tasks involves constructing a risk assessment tool for grant reviews, which in real-world professional practice requires formal sign-off and archival, placing sign-off logging and archival compatibility into this category. When attributing the origin of a requirement, priority is given first to task materials, then to benchmark claims; properties are classified as analytic inference only when neither explicitly mandates them.

We define an evaluation gap strictly as an instance where a property is task-specified or claim-specified but left uninspected by the evaluated result. Requirements identified through analytic inference define a scope boundary instead. Because the benchmark did not intend to evaluate these properties, its scores remain valid with respect to its stated claims; the boundary serves only to delimit the broader work-related conclusions that the scores can warrant \citep{kane2013validating}. Documenting the provenance of each requirement ensures that judgments regarding evaluation gaps can be systematically audited, enabling secondary annotators to inspect the underlying instructions or claims.

\subsection{Evaluated Result}
\label{sec:scored-object}

The evaluated result field records which components of the required work product are actually examined by the benchmark. Reporting must enumerate every evaluation instrument released alongside the tasks (e.g., executable tests, rubrics, and expert comparisons) and specify the precise properties assessed by each. Benchmarks that assess only a final answer or a binary success state leave the remainder of the work product uninspected; likewise, multi-step executions involving web navigation or tool use are often ultimately scored against a brief final answer. This field focuses exclusively on what is inspected; measurement quality concerns, such as sample size and inter-annotator agreement, are outside the scope of this paper.

In the document revision example, evaluating only a rewritten passage indicates that the system generated revised text, but reveals little about the integrity of the full draft. Evaluating the revised draft itself requires verifying that source-grounded content is preserved and that reviewers can trace the basis for each modification. Fluency and localized improvements represent a narrower objective than a fully grounded, traceable revision.

The target properties under evaluation derive directly from the preceding two fields: the tested setting determines whether compliance with designated source materials should be assessed, while the required work product dictates whether the artifact preserves what downstream stakeholders require. Generic criteria such as correctness remain valuable, yet broad, monolithic judgments risk conflating ungrounded evidence usage with overall content quality. When rubrics are employed, benchmark reports should explicitly trace how each criterion derives from the required work product and describe how annotators apply it in practice \citep{mislevy2003structure}.

\begin{figure}[!ht]
\centering
\includegraphics[width=\textwidth]{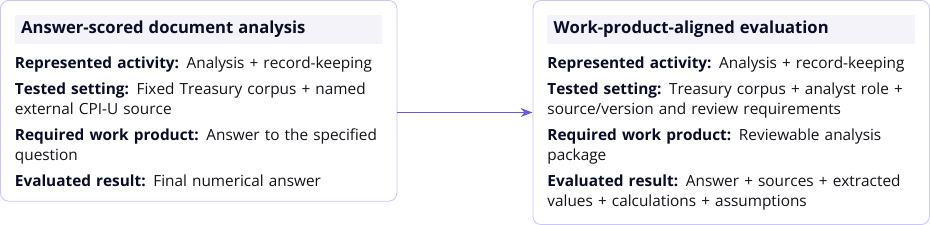}
\caption{Same represented activity, different work-product requirements and evaluation targets. The left panel shows an answer-scored document-analysis task of the form examined in OfficeQA Pro. When the intended product is a reviewable analyst deliverable, source/version information, extracted values, calculations, and assumptions also enter the evaluation.}
\label{fig:officeqa-claim}
\end{figure}

These four fields are documented independently because changes in one do not necessarily alter the others. In Figure~\ref{fig:officeqa-claim}, two tasks share the same represented activity yet require different work products, necessitating different checks during evaluation. When designing new benchmarks, creators can populate the fields sequentially---from activity to tested setting, followed by work product---before determining which elements of the work product to evaluate. Conversely, when analyzing existing benchmarks, researchers can work backwards from public artifacts to identify gaps left uninspected by final scores, as demonstrated in Section~\ref{sec:whybench}.

\subsection{A Shared Work Activity Vocabulary}
\label{sec:vocab}

The \emph{represented activity} field can be populated using different activity vocabularies. O{*}NET itself provides Generalized, Intermediate, and Detailed Work Activities \citep{onet2026overview}. Among these, the intermediate and detailed levels are too granular to support consistent benchmark labeling, whereas the generalized level organizes work at a hierarchy that diverges from our representation. We therefore start directly from O{*}NET task statements \citep{onet2024tasks} and group them by the transformations performed and the artifacts produced in the work, deriving an 18-category cross-occupational vocabulary for benchmark reporting, which we adopt in Section~\ref{sec:whybench}. Other objectives, such as labor market analysis, may call for a different vocabulary \citep{acemoglu2011skills}.

\begin{figure}[!htbp]
\centering
\includegraphics[width=\textwidth]{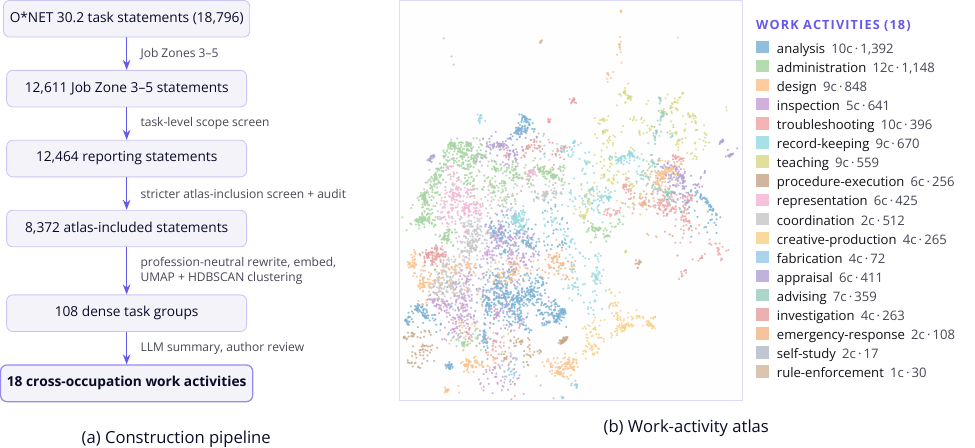}
\caption{Construction and resulting atlas of the 18-work-activity inventory. Panel (a) separates the reporting corpus from the stricter atlas subset. Panel (b) shows the atlas used to inspect the work-activity grouping.}
\label{fig:inventory}
\end{figure}

Figure~\ref{fig:inventory} outlines the construction process and the resulting taxonomy. The derivation begins with the 18,796 task statements in O{*}NET 30.2. Restricting the candidate occupations to Job Zones 3--5 (occupations requiring medium to extensive preparation) leaves 12,611 statements. Task-level scope filtering removes 147 statements whose primary content falls outside our reporting scope, leaving 12,464. A stricter taxonomy inclusion filter initially excludes 4,126 statements; following an author audit, 34 statements regarding scheduling availability are added back, resulting in a final set of 8,372 statements used for clustering (Appendix~\ref{app:taxonomy}). The final label set was determined through human review and subsequently applied to the 12,464-statement reporting corpus. Appendix~\ref{app:taxonomy} reports four validation checks on the finalized vocabulary. Two independent reviewers recognized most labels without definitions and correctly matched all 18 once definitions were provided; the checks also revealed that the boundary between procedure-execution and record-keeping is relatively weak.

\FloatBarrier

Table~\ref{tab:ops} lists the frequency, definitions, example scenarios, and common benchmark proxy tasks across the 18 work activities in our reporting corpus. The proxy tasks column corresponds to the component labels described in Section~\ref{sec:define-work}, which typically cover only a fraction of any given activity. When a more fine-grained distinction than these 18 categories is required, researchers can refer back to the 108 original clusters, each mapping to a single label (Appendix~\ref{app:taxonomy}).

\begin{table}[!ht]
\caption{The 18 cross-occupation work activities derived from O{*}NET task statements. Counts are final activity assignments across the 12{,}464-task reporting corpus; the atlas subset contains 8{,}372 statements, so reporting and atlas counts differ by activity. The counts describe the reporting vocabulary and carry no labor-market weighting or AI deployment estimate. Occupations/settings and benchmark proxies are illustrative.}
\label{tab:ops}
\centering
\scriptsize
\setlength{\tabcolsep}{3pt}
\renewcommand{\arraystretch}{1.08}
\begin{tabularx}{\linewidth}{
@{}
>{\ttfamily\arraybackslash}p{0.16\linewidth}
>{\centering\arraybackslash}p{0.055\linewidth}
>{\raggedright\arraybackslash}p{0.32\linewidth}
>{\raggedright\arraybackslash}p{0.2\linewidth}
>{\raggedright\arraybackslash}X
@{}}
\toprule
\textbf{\normalfont Work activity} &
\textbf{\normalfont Tasks} &
\textbf{\normalfont Definition} &
\textbf{\normalfont Example settings} &
\textbf{\normalfont Common proxies} \\
\midrule
analysis &
1,732 &
Synthesize evidence into a supported finding or interpretation. &
Research; finance. &
Retrieval; summarization. \\
administration &
1,517 &
Plan, allocate, or maintain organizational resources or processes. &
Program management; operations. &
Form filling; state tracking. \\
design &
992 &
Specify a system, plan, artifact, or service to satisfy a goal. &
Product design; architecture. &
Plan generation; specification writing. \\
inspection &
963 &
Check a current state against rules, standards, or specifications. &
Compliance; safety review. &
Classification; checklist scoring. \\
troubleshooting &
902 &
Diagnose a problem and apply or recommend a remedy. &
IT support; software debugging. &
Log analysis; state repair. \\
record-keeping &
882 &
Create, update, store, or retrieve canonical organizational records. &
Clinical records; case files. &
Extraction; database update. \\
teaching &
814 &
Support a learner in acquiring knowledge or skill. &
Tutoring; professional training. &
Explanation; feedback generation. \\
procedure-execution &
735 &
Carry out a prescribed procedure according to a standard. &
Laboratory workflow; service procedure. &
Policy following; checklist execution. \\
representation &
615 &
Speak or act on behalf of an organization, client, or role. &
Client communication; advocacy. &
Role-play dialogue; policy-constrained response. \\
coordination &
603 &
Align actors, constraints, dependencies, or timing around shared work. &
Care coordination; logistics. &
Scheduling; task routing. \\
creative-production &
543 &
Produce an original creative, media, or communicative artifact to a brief. &
Marketing; media production. &
Copywriting; slide generation. \\
fabrication &
505 &
Plan or supervise the production, installation, or maintenance of a tangible asset from a work specification. &
Construction planning; equipment setup. &
Instruction generation; plan checking. \\
appraisal &
453 &
Assess, grade, value, or judge an artifact or case against a standard. &
Peer review; claims review. &
Rubric scoring; preference judgment. \\
advising &
403 &
Tailor a recommendation to a specific person, client, or case. &
Clinical advice; financial advice. &
Chat response; recommendation generation. \\
investigation &
377 &
Reconstruct an event, condition, or pattern from evidence. &
Fraud analysis; epidemiology. &
Timeline reconstruction; evidence retrieval. \\
emergency-response &
212 &
Stabilize or triage an acute incident under time pressure. &
Clinical triage; incident response. &
Triage classification; escalation decision. \\
self-study &
186 &
Maintain or update one's working competency through learning. &
Certification; continuing education. &
Learning plan; self-assessment. \\
rule-enforcement &
30 &
Apply or uphold behavioral or procedural rules in a setting. &
Conduct review; safety enforcement. &
Violation classification; escalation. \\
\bottomrule
\end{tabularx}
\end{table}

\section{Benchmark Case Analyses Through the Work-Centered Representation}
\label{sec:whybench}

In this section, we apply the representation introduced in Section~\ref{sec:guidelines} to three public benchmark cases.\footnote{Each case reflects a distinct evaluative architecture, ranging from expert-reviewed deliverables to executable state transitions.} For each case, we populate the four fields in Table~\ref{tab:case-comparison}, identify requirements within the required work product that are left unchecked, and follow the decision rules in Section~\ref{sec:score-work-product} to classify each omission as either an evaluation gap or a scope boundary. We use a single public instance from each benchmark for case-level demonstration; for APEX-SWE, we further extend our analysis to an audit of the tests across all 25 public Integration tasks. For GDPval and OfficeQA Pro, achieving benchmark-level coverage would require stratified sampling within each suite. A second author coded the four fields for the three cases using only public materials, without access to Table~\ref{tab:case-comparison}. Across the 12 fields, the two codings were in full agreement on 7 fields and in partial agreement on the remaining 5; the partial agreements involved two activity labels and three descriptive refinements caught by the second coder, all of which are reflected in Table~\ref{tab:case-comparison}. Following this first coding round, we clarified our definitions (Appendix~\ref{app:audit}). A third author then independently recoded the three cases without access to Table~\ref{tab:case-comparison} or the prior two codings, achieving full agreement with Table~\ref{tab:case-comparison} on 10 of the 12 fields; the two partial agreements involved an activity label in GDPval and a named source in the GDPval prompt, both of which are incorporated into Table~\ref{tab:case-comparison}.

\begin{table*}[!ht]
\caption{The four fields populated across three public cases. The table reflects the case-level interpretations used in the subsequent analyses.}
\label{tab:case-comparison}
\centering
\footnotesize
\setlength{\tabcolsep}{4pt}
\renewcommand{\arraystretch}{1.12}
\begin{tabularx}{\textwidth}{@{}>{\raggedright\arraybackslash}p{0.14\textwidth}>{\raggedright\arraybackslash}X>{\raggedright\arraybackslash}X>{\raggedright\arraybackslash}X@{}}
\toprule
\textbf{Field} & \textbf{GDPval} & \textbf{OfficeQA Pro} & \textbf{APEX-SWE} \\
\midrule
\textbf{Represented activity} & design. & record-keeping + analysis. & procedure-execution + record-keeping; troubleshooting added upon execution failure. \\
\textbf{Tested setting} & Federal grant management role, no reference documents provided (2 C.F.R. Part 200 named in prompt as external source), task context, and specified deliverables. & Fixed Treasury Bulletin corpus, one named external CPI-U source, and closed numerical questions. & Local \texttt{users.csv}, credentials, service documentation, LocalStack state, executable tests, and rubric. \\
\textbf{Required work product} & Federal Applicant Risk Assessment Tool tailored to a specified grant review task. & Numerical answer to the specified question, computed from the correct Bulletin figures and the specified CPI-U value. & A runnable Python script alongside the specified S3 objects and state changes. \\
\textbf{Evaluated result} & Submitted compliance deliverable judged via pairwise expert comparison; public task rubric checks formatting, headings, length, topic coverage, and citations. & Final numerical value evaluated via exact match or numerical tolerance. & Execution, S3 paths, schema, and non-empty outputs; line-by-line correspondence with source file left unchecked. \\
\textbf{Unchecked requirement (source)} & Integration into grant workflow, approval records, and audit trail (analytic inference; scope boundary). Under the public rubric, universal applicability across applicant types and conflict-of-interest exceptions (task-specified; evaluation gap). & Intermediate monthly totals and specified CPI-U value (task-specified), as well as grounding in cited Bulletins (claim-specified); all evaluation gaps. & Line-by-line correspondence between \texttt{users.csv} and uploaded objects (task-specified; evaluation gap). \\
\bottomrule
\end{tabularx}
\end{table*}

GDPval is introduced as a benchmark evaluating model capabilities on realistic, economically valuable knowledge-work tasks across 44 occupations, releasing 5 gold cases per occupation. It requires deliverables such as documents and spreadsheets, scoring outputs via pairwise expert comparisons between model deliverables and human expert deliverables \citep{gdpval2025}. In one public compliance case (task id \texttt{36d567ba-\allowbreak e205-\allowbreak 4313-\allowbreak 9756-\allowbreak 931c6e4691fe}), the model is placed in a federal grant management role and tasked with producing a Federal Applicant Risk Assessment Tool for grant reviews; this tool constitutes the required work product. The represented activity is design. Inspection and record-keeping enter only downstream when the tool is utilized during actual reviews, whereas the public task itself terminates at deliverable submission. The tested setting specifies the role and deliverable format; the public materials provide no reference documents, and the prompt merely names 2 C.F.R. Part 200 (the Uniform Guidance) as an external source. The evaluated result stems from two public grading instruments: expert comparisons evaluating the compliance deliverable alongside human expert work, and a public 54-point rubric specific to the task. The rubric verifies presentation style and Uniform Guidance citations, but fails to check the prompt requirement that the tool apply across all applicant types; similarly, while it mandates a yes/no question for every topic, it overlooks the prompt's explicit exception for conflicts of interest.

In GDPval, a self-contained digital deliverable serves as a task proxy for a slice of occupational work. This setting isolates computer-based labor and predetermines task boundaries. Evaluation terminates at deliverable quality, leaving the downstream grant review workflow that utilizes the tool outside the public case. Because only 5 cases are public per occupation, occupation-level coverage requires additional sampling from tasks to activities.

OfficeQA Pro is introduced as a benchmark for grounded multi-document reasoning over an extensive corpus of U.S. Treasury Bulletins. Its tasks require retrieval and synthesis across text and tables, with the Pro subset featuring challenging questions paired with verifiable answers \citep{opsahlong2026officeqapro}. The national defense expenditures case (task id \texttt{UID0005}) requires the model to combine monthly figures reported in 1953 and 1940 with an external CPI-U figure to execute an inflation-adjusted comparison. Under the work-centered representation, this case embodies record-keeping and analysis, as the model must first locate the correct Treasury Bulletin records before computing with them. The tested setting consists of a fixed document corpus, a closed analytical question, and an external CPI-U source named in the prompt but not provided directly. The required work product is a numerical answer to the question, computed from the correct Bulletin figures and the specified CPI-U value, and the evaluated result is the final value judged by exact match or numerical tolerance. Closed questions make numerical scoring reproducible, but evaluation stops strictly at the final number. A reviewable analyst work product would additionally include cited sources and intermediate derivations, both of which would be verified under operational evaluation.

APEX-SWE is introduced as a benchmark of economically valuable software engineering tasks. Integration tasks require building end-to-end systems across cloud and business services, while Observability tasks require debugging production-like outages. Public tasks provide containerized environments complete with pytest suites and rubric files \citep{kottamasu2026apex}. In the \texttt{1-\allowbreak aws-\allowbreak s3-\allowbreak snapshots} case, the model must write a Python script that uploads a local \texttt{users.csv} file to a date-stamped path in LocalStack S3. The represented activity primarily comprises procedure-execution and record-keeping, with troubleshooting engaged when execution failures arise in the benchmark environment. The tested setting supplies local data, an active running service, and executable tests. The required work product is a runnable script alongside the specified S3 state changes. The test suite checks that the script runs and writes a correctly formatted, non-empty file to the target path. The public reference solution reads the local \texttt{users.csv} as instructed, but the tests never check line-by-line correspondence between the source file and the uploaded object \citep{mercor2026apexs3}. While the public rubric lists reading \texttt{users.csv} as a functional criterion, the evaluation harness scores pass/fail exclusively on test execution, and neither the rubric nor the tests cross-check uploaded rows against source records.

In this APEX-SWE case, the tests confirm local task execution and end-state arrival, leaving consistency from source to state only partially checked. Production deployment lies outside the public integration episode. Supporting conclusions about source data fidelity requires a check verifying uploaded records against the designated source file.

\paragraph{What changes after the record is filled.}
Evaluation terminates at different stages of work across the three cases: GDPval stops at deliverable quality, OfficeQA Pro stops at the final numerical value, and APEX-SWE stops at end-state presence. In APEX-SWE, correspondence between uploaded rows and \texttt{users.csv} is never checked; because uploading the local \texttt{users.csv} is explicitly mandated, this omission is a task-specified evaluation gap. A pass on \texttt{1-\allowbreak aws-\allowbreak s3-\allowbreak snapshots} supports the conclusion that the script ran and deposited a correctly formatted, non-empty object at the specified path; it cannot support the conclusion that the designated records were actually transferred. Running the public reference solution and three modified variants against the 6 public tests without alteration demonstrates this gap: one script uploads 5 dummy users per the internal specification doc rather than the agent-facing issue, one uploads only the first row, and one scrambles all email addresses. All four scripts pass all 6 tests. Adding a single assertion checking uploaded rows against \texttt{users.csv} passes the reference solution while failing all three modified scripts, requiring no changes to the task definition itself.

In OfficeQA Pro, the evaluator compares only the final number against reference values within benchmark tolerance. Points are awarded whenever values match---regardless of derivation method (e.g., parametric memorization)---and intermediate monthly totals or specified CPI-U figures are never scored; hence, a correct final value does not itself constitute evidence of grounded reasoning. The intermediate monthly totals and specified CPI-U figures are task-specified requirements, whereas grounding in cited Bulletins stems from the benchmark's claim of grounded reasoning (claim-specified); both omissions represent evaluation gaps. Requiring cited tables and extracted values alongside the final answer would allow the benchmark to decouple raw answer accuracy from grounded answer accuracy, where only the latter supports claims regarding grounding.

In GDPval, pairwise expert comparisons evaluate the deliverable holistically; unchecked properties external to the deliverable (e.g., workflow continuity, approval sign-offs, and audit trails) are neither provided in task materials nor claimed by the benchmark. These represent analytic inferences and mark a scope boundary. Consequently, the benchmark itself requires no modification; rather, claims must be bounded to expert-judged deliverable quality on a discrete task, rather than competence across an entire grant review workflow. However, the public rubric omits two task-specified requirements (universal applicant applicability and the conflict-of-interest exception), introducing evaluation gaps that render rubric-based scores narrower in support than expert comparisons until resolved.

\paragraph{Test audit of APEX-SWE Integration tasks.}
To verify whether our APEX-SWE findings extend beyond individual cases, we coded the public test suites across all 25 Integration tasks. For each task, the agent-facing issue defines a primary deliverable alongside a source governing its contents (e.g., a provided data file or seeded service records). We coded each test suite as follows: \emph{full} if every source-determined element in the deliverable is validated against its source; \emph{partial} if some source-determined elements are checked while others are ignored; and \emph{none} if the tests verify only source-independent properties (e.g., existence, execution, destination path, formatting schema, valid ranges, or internal consistency between outputs of the same script). Coding was conducted via manual inspection of public test code; for \texttt{1-\allowbreak aws-\allowbreak s3-\allowbreak snapshots} and \texttt{283-\allowbreak zammad-\allowbreak sla-\allowbreak archive}, we also executed empirical counterexamples. A second author independently coded all 25 tasks under identical definitions without access to the first round of codes. The two codings agreed on 21 of 25 levels (Cohen's $\kappa =0.73$). The 4 discrepancies (detailed in Appendix~\ref{app:audit}) were adjudicated against the written definitions, adopting the second author's coding in 3 instances and the first author's in 1.

We subsequently refined our definitions to address edge cases underlying these disagreements (Appendix~\ref{app:audit}). A third author then coded all 25 tasks under the revised definitions without access to prior codings, achieving perfect agreement with the adjudicated set across all 25 levels (Cohen's $\kappa =1.00$); these tasks constituted the exact set over which definitions were clarified. In the adjudicated coding, 12 tasks were classified as \emph{full}, 10 as \emph{partial}, and 3 as \emph{none}. Thus, in 13 of the 25 tasks, deliverables can mutate source-governed contents while still passing every public test (judged from the test code; confirmed by execution in two of them).

Unchecked properties concentrate heavily in aggregate outputs. 10 of the 12 \emph{full} tasks produce single records or discrete scalar values (e.g., syncing a single ticket to an external platform or writing a generated secret to file), whereas 11 of the 13 \emph{partial} or \emph{none} tasks generate multi-record collections. In \texttt{283-\allowbreak zammad-\allowbreak sla-\allowbreak archive}, a script that never queries the Zammad ticketing system, uploads an empty CSV containing only header rows, and reports zero archived tickets passes all 14 public tests. In \texttt{247-\allowbreak customer-\allowbreak journey-\allowbreak analytics}, the test verifies only that health scores align internally with the report's self-reported counts, without checking those counts against underlying source event records. Furthermore, 5 of the \emph{full} tasks output terminal confirmation strings (e.g., ticket IDs or counts) that remain unasserted by the tests; these confirmation strings are documented in Appendix~\ref{app:audit} and excluded from the summary counts above.

\paragraph{What the activity field adds.}
Across APEX-SWE and OfficeQA Pro, the evaluated result covers only one activity within the episode, leaving the other unchecked; in GDPval, the unchecked activities fall outside task boundaries altogether. In APEX-SWE, the tests verify procedure-execution (verifying that the script runs and targets the correct location) while omitting record-keeping (verifying that uploaded rows correspond to source records). In OfficeQA Pro, the harness evaluates the numerical output of analysis, bypassing the record-keeping step of retrieving the correct Bulletin figures. In GDPval, expert comparisons evaluate the design product, while downstream inspection and record-keeping reside outside the task.

Appendix~\ref{app:opslens} outlines the evidentiary properties each activity product must preserve. For record-keeping, it specifies provenance and field-level integrity, precisely the dimensions left unchecked across 13 of the 25 audited APEX-SWE tasks, most notably in multi-record deliverables. Tagging an episode with its represented activity thus alerts benchmark designers to essential product properties prior to running an audit. Symmetrically, these labels enable comparative audit syntheses across disparate benchmarks along shared activity lines.

\paragraph{Artifacts and reproducibility.}
The supplementary materials include machine-readable definitions for the 18 work activities, alongside validation summaries corresponding to Appendix~\ref{app:taxonomy}. Full O{*}NET assignment outputs will be released alongside the published version, and Appendix~\ref{app:taxonomy} details the taxonomy construction process. During our audit in October 2026, the official APEX-SWE repository returned access errors; we recovered the benchmark tasks from a fixed revision on a public mirror.\footnote{\raggedright Official release: \url{https://huggingface.co/datasets/mercor/APEX-SWE} (returned HTTP 401/404 on 2026-10-19 and 2026-10-20). Public mirror utilized: \url{https://www.modelscope.cn/datasets/mercor/APEX-SWE}, revision \texttt{ef715ef646ec8d681dd200b538e02bdb908cf06f}. Evaluation pass/fail logic is executed by \texttt{integration/src/harness/evaluator.py} in \url{https://github.com/Mercor-Intelligence/apex-swe}.} The supplementary materials also include the audit codes from Appendix~\ref{app:audit} and the counterexample scripts, with corresponding task files accessible via the mirrored revision.

\section{Discussion}
\label{sec:altviews}

When a benchmark score is interpreted as evidence of work competence, claims must match the granularity of the underlying evidence. Direct workplace studies answer localized, concrete questions without relying on intermediary benchmarks. Field deployments and randomized experiments capture end-to-end productivity metrics across specific worker cohorts and established workflows \citep{dellacqua2023jagged,brynjolfsson2023genai}. When cohorts and workflows align with targeted deployments, these studies directly warrant local conclusions, though generalizability to external organizations demands separate empirical support. Public benchmarks serve a distinct role: they provide reproducible, pre-deployment comparative evaluations across systems. Reporting the four fields makes the operational boundaries of this evidence explicit.

Representative work samples and predictive proxies similarly support distinct claims. Benchmarks sampled from explicit job or workflow taxonomies and evaluated on artifacts or outcomes generated in situ can directly evaluate target domains \citep{roth2005worksample}. When simpler capability metrics exhibit proven predictive validity within a target domain, they reliably support model rankings or performance forecasts \citep{ruan2024observational}. Portfolio-level taxonomies can further trace where evaluation effort concentrates across sectors \citep{wang2026realwork}. Making broader claims about work competence, however, requires explicitly reporting the required work product alongside the evaluated result. When claims are bounded strictly to component execution, component evaluations suffice. Retrieval and embedding benchmarks intentionally isolate modular behaviors, making failure modes straightforward to diagnose \citep{thakur2021beir,muennighoff2022mteb}. In contrast, broader conclusions regarding a composite activity (such as analysis) depend heavily on how components operate within the tested setting and what artifacts downstream workflows mandate.

We employ the O{*}NET inventory as one operational vocabulary for the represented activity field. Work practices naturally diverge across institutions and evolving temporal contexts \citep{suchman1987plans}, and O{*}NET task statements capture occupational labor at a high level of abstraction, inheriting the inherent constraints of occupational classification systems \citep{handel2016onet,onet2024tasks}. As more granular workflow records emerge, this taxonomy can be refined or superseded. Its function here is to offer a standardized, mid-level vocabulary spanning the divide between granular component tasks and aggregate occupational titles; in Section~\ref{sec:whybench}, these categorical labels directly expose unchecked product properties across benchmarks.

Benchmark evaluations can likewise expand from terminal metrics to encompass broader segments of the required work product. Software engineering benchmarks already utilize executable test suites to evaluate persistent repository state \citep{jimenez2024swebench,kottamasu2026apex}, just as web and desktop agent benchmarks expose state transitions across environments \citep{zhou2023webarena,xie2024osworld,drouin2024workarena}. For clinical or scientific tasks, this design paradigm applies symmetrically: first specify the represented activity and tested setting, delineate the exact deliverables handed off to downstream users or systems, and verify the specific components necessary to support intended claims. Because knowledge-work benchmark results increasingly inform workplace automation policies, interpretive claims must remain strictly anchored to what was empirically measured. Macro-level assertions regarding labor displacement require occupational and workplace-level evidence.

The work activity taxonomy itself warrants broader empirical validation. Comparing these labels against observed enterprise workflows and workplace trace logs will determine their utility outside the O{*}NET framework \citep{handa2025economic}. The open-label interpretability evaluation in Appendix~\ref{app:taxonomy} was conducted by two independent reviewers across 18 blind task cards; recruiting larger, independent panels to evaluate wider task samples will clarify whether observed ambiguities (such as classifying procedure-execution as navigation) persist across broader contexts.

The analytical methodology applied to APEX-SWE can be extended to GDPval and OfficeQA Pro through stratified, multi-instance sampling coupled with independent coding and formal adjudication. Because our APEX-SWE audit targeted Integration tasks, evaluating Observability tasks---which center on diagnosing production failures---will require distinct formal criteria for what constitutes source-determined state. Finally, empirical studies can directly test the utility of the framework itself by comparing design choices across benchmark developers with and without four-field reporting. These inquiries will help establish both the coverage boundaries of existing benchmarks and the practical utility of the representation.

\section{Conclusion}
\label{sec:conclusion}

Knowledge work varies fundamentally both in what actions are performed and in what artifacts must persist upon completion. In this paper, we propose a work-centered benchmark representation structured around four fields: represented activity, tested setting, required work product, and evaluated result. Across three public benchmark cases, evaluated work terminates at distinct stages of execution, yet all map into this shared representation; the resulting profiles identify precisely which claims current scores support and what evaluations must be added to warrant broader conclusions. Benchmark design can begin by anchoring to a work activity and tested setting, specifying the work product that activity must yield, and implementing evaluators that inspect the necessary facets of that deliverable. Explicitly reporting these elements clarifies the boundaries of claims whenever agents are evaluated in work-like settings.

\newpage
\renewcommand{\bibfont}{\normalsize}
\bibliographystyle{plainnat}
\bibliography{references}

\newpage
\appendix
\renewcommand{\thefigure}{\Alph{section}.\arabic{figure}}
\renewcommand{\thetable}{\Alph{section}.\arabic{table}}
\makeatletter
\@addtoreset{figure}{section}
\@addtoreset{table}{section}
\makeatother
\setcounter{figure}{0}
\setcounter{table}{0}

\section{Inventory Construction Details}
\label{app:taxonomy}

\paragraph{Corpus.}
This appendix details the construction of the 18-class work activity taxonomy used in Section~\ref{sec:vocab} from O{*}NET task statements \citep{onet2024tasks}. Figure~\ref{fig:inventory}(a) illustrates the full pipeline, including the clustering stage yielding 108 dense task groupings prior to manual consolidation \citep{mcinnes2018umap,campello2013density}. Beginning with 18,796 task statements from O{*}NET 30.2, we restricted the candidate pool to occupations within Job Zones 3--5, yielding 12,611 statements \citep{onet2024tasks}. The Job Zone threshold establishes the target occupational boundary. An iterative screening process supported by GPT-5.5 and adjudicated by the authors removed 147 statements centered predominantly on direct physical labor, performance, or routine clerical duties, leaving 12,464 task statements as the reporting corpus. Each retained statement preserves its O{*}NET identifier, ratings, and Job Zone designation.

\paragraph{Screens.}
The screening procedure combined iterative GPT-5.5 classification with author adjudication. We drew three non-replacement batches of 100 tasks from the candidate corpus, evaluating each batch against an expanding list of candidate exclusion categories until the list stabilized at 12 classes. Screening explicitly targeted physical execution, performing arts, and routine clerical work; tasks involving planning, specification, supervision, or inspection of tangible labor were retained provided their outputs were cognitive or documented in nature.

We then applied a stricter graph-inclusion screen to the 12,464 statements. This filter initially excluded 4,126 items; author review reinstated 34 statements pertaining to scheduling availability, resulting in 8,372 statements for clustering and consolidation. The final 18 work activity labels were applied back across the full 12,464 reporting statements to produce Table~\ref{tab:ops}; the graph visualizes the tighter 8,372-statement subset.

\paragraph{Embedding and clustering.}
Task statements were rewritten under a fixed prompt into decontextualized work activity phrases to eliminate domain-specific vocabulary. These were embedded into a 1,536-dimensional space using \texttt{text-embedding-3-small} and $L_{2}$-normalized. Dimensionality reduction via UMAP (15 dimensions, 30 neighbors, cosine distance, min\_dist 0, random seed 42) was followed by HDBSCAN (min\_cluster\_size 30, min\_samples 10, excess-of-mass extraction), identifying 108 clusters purely from task geometry; 4,487 points designated as noise were softly assigned via membership vectors for subsequent review \citep{mcinnes2018umap,campello2013density}.

\paragraph{Consolidation.}
The authors examined all 108 raw clusters, merging or partitioning groups based on operational transformations and preserved work products. Four rounds of summarization via Claude Opus 4.7 (temperature 0.1) proposed decontextualized cluster labels, with final inclusions determined by expert review. Rounds 1--3 yielded 38, 39, and 16 top-level activities, respectively, none of which were retained. Round 4 produced 19 candidate activities, maintaining inspection and investigation as distinct categories. Manual review merged a 93-task clinical, surgical, and nursing support cluster into procedure-execution, finalizing the 18 work activities shown in Figure~\ref{fig:inventory} and Table~\ref{tab:ops}.

\paragraph{Assignment checks.}
Every cluster in the graph possesses a single top-level assignment; each top-level work activity contains at least one constituent cluster and spans diverse occupational families across its exemplar set. For reporting, final labels were mapped across all 12,464 screened task statements, summing to the complete reporting corpus in Table~\ref{tab:ops}. Expansion ratios from graph subsets to reported assignments vary widely across activities: self-study expanded from 17 graph tasks to 186 reported statements (10.9$\times$), fabrication expanded from 72 to 505 (7.0$\times$), and rule-enforcement remained unchanged at 30 to 30 (1.00$\times$). These ratios demarcate assignment boundaries rather than labor market shares. Each activity is defined alongside explicit boundaries distinguishing it from adjacent categories; for example, self-study denotes active skill maintenance or acquisition, excluding mandatory attendance requirements.

\paragraph{Validation.}
Table~\ref{tab:inventory-validation} summarizes four validation checks on the finalized taxonomy. Two independent reviewers identified most activity labels without access to definitions, and matched all 18 correctly once definitions were revealed. Reviewer feedback indicates that the boundary between procedure-execution and record-keeping is the most permeable.

\begin{table*}[!ht]
\caption{Four validation questions for the 18-class work activity reporting taxonomy, alongside evidentiary results, interpretations, and remaining limitations.}
\label{tab:inventory-validation}
\centering
\footnotesize
\setlength{\tabcolsep}{4pt}
\renewcommand{\arraystretch}{1.12}
\begin{tabularx}{\textwidth}{@{}>{\raggedright\arraybackslash}p{0.14\textwidth}>{\raggedright\arraybackslash}p{0.34\textwidth}>{\raggedright\arraybackslash}p{0.23\textwidth}>{\raggedright\arraybackslash}X@{}}
\toprule
\textbf{Question} & \textbf{Evidence and Results} & \textbf{Interpretation} & \textbf{Limitations} \\
\midrule
\textbf{Internal consistency} & Cosine similarity from tasks to activity centroids, alongside blind human consistency ratings (1--5 scale). Median similarities across activities ranged from 0.538 to 0.911 (weighted mean 0.581). Mean human ratings were 3.833 and 4.056 (median 4.0 for both). & Tasks within a given activity exhibit semantic coherence across both embedding similarity and human review. & Descriptive similarity metrics cannot establish whether the clustering solution is mathematically unique. \\
\textbf{Algorithmic sensitivity} & Comparison between HDBSCAN non-noise assignments and KMeans ($k=40$). Adjusted Rand Index = 0.460; Normalized Mutual Information = 0.822. & Broad groupings are partially preserved under KMeans partitioning. & Random seed sensitivity was unmeasured; $k=40$ imposes an alternate granularity. \\
\textbf{Open-label interpretability} & Two independent reviewers evaluated 18 blind cards in randomized order, providing open labels, consistency scores, and outlier flags. Consistency scores agreed identically on 12/18 cards and within $\pm 1$ point on 18/18 (quadratic weighted $\kappa =0.802$). Direct recognition reached 15/18 and 16/18; direct or partial recognition reached 17/18 for both; 3 common outliers emerged across 71 exemplar tasks. & Most activities are recognizable from task exemplars alone without prior definitions. & Limited to two reviewers, hand-curated cards, author-coded correspondences, and a partially annotated corpus. \\
\textbf{Definition matching} & Secondary matching phase following definition disclosure. Both reviewers matched 18/18 correctly. & Reviewers cleanly distinguish all 18 categories once definitions are provided. & Confirms definitional clarity; procedure-execution remains the most difficult class to identify in open-label settings. \\
\bottomrule
\end{tabularx}
\end{table*}

\section{Design Notes for Each Work Activity}
\label{app:opslens}

Table~\ref{tab:opslens} provides a representative benchmark proxy task and corresponding evidentiary work product requirements for each of the 18 work activities.

\begin{table}[!ht]
\caption{Design considerations when applying the work-centered representation across work activities.}
\label{tab:opslens}
\centering
\scriptsize
\begin{tabularx}{\linewidth}{@{}>{\raggedright\arraybackslash}p{0.17\linewidth}>{\raggedright\arraybackslash}X>{\raggedright\arraybackslash}X@{}}
\toprule
\textbf{Work activity} & \textbf{Viable benchmark proxy task} & \textbf{Required evidence in evaluated work product} \\
\midrule
\textbf{analysis} & Interpretive briefing memo synthesizing heterogeneous evidence. & Must preserve source attribution, underlying assumptions, inferential reasoning, and verifiable conclusions. \\
\textbf{administration} & Multi-system workflow task spanning records, approvals, and enterprise tools. & Must capture updated state, authorization rationale for changes, and remaining handoff or approval milestones. \\
\textbf{design} & Technical specification for a system, service, deliverable, or operational workflow. & Must be actionable for downstream implementers, detailing constraints, trade-offs, and acceptance criteria. \\
\textbf{inspection} & Systematic audit of an artifact against rules, standards, or contractual clauses. & Must cite governing standards, identify specific inspection targets, and delineate compliant vs. non-compliant findings. \\
\textbf{troubleshooting} & Diagnostic and remediation task in software, clinical, mechanical, or operational settings. & Must include root-cause diagnosis, evidentiary support, proposed fixes, and verification of state resolution. \\
\textbf{record-keeping} & Retrieval, updating, reconciliation, or archiving of an authoritative record. & Must preserve versioning, provenance, field-level mutations, retention constraints, and retrieval pathways. \\
\textbf{teaching} & Instructional vignette or feedback module targeted to a learner. & Must articulate learning objectives, learner state, instructional interventions, and evidence of pedagogical utility. \\
\textbf{procedure-execution} & Deterministic protocol execution following established operating standards. & Must document mandatory steps executed, deviations, escalation choices, and terminal process state. \\
\textbf{representation} & Formal communication or negotiation conducted on behalf of an entity or role. & Must capture role authorization, constraints, commitments made, and external records of engagement. \\
\textbf{coordination} & Multi-tool synchronization of project stakeholders, timelines, dependencies, or handoffs. & Must articulate ownership, dependency paths, outstanding blockers, and designated downstream handoffs. \\
\textbf{creative-production} & Production of media, creative designs, or strategic messaging against a creative brief. & Must fulfill creative requirements, incorporate mandated assets or constraints, and export to specified formats. \\
\textbf{fabrication} & Planning or supervision of physical build processes against technical specifications. & Must capture fabrication sequencing, material/safety constraints, acceptance criteria, and physical proxy limits. \\
\textbf{appraisal} & Formal valuation, rating, quality audit, or evaluative assessment of a case or asset. & Must articulate applied rubrics, inspected evidence, evaluative judgments, and justifications verifiable by a secondary auditor. \\
\textbf{advising} & Tailored advisory recommendations delivered to a client, patient, or stakeholder. & Must reflect case facts, considered alternatives, scope constraints, risk/uncertainty disclosures, and actionable guidance. \\
\textbf{investigation} & Reconstruction of an incident, anomaly, or temporal pattern from fragmented evidence. & Must detail evidentiary discovery, source conflicts, causal or chronological timelines, and unresolved gaps. \\
\textbf{emergency-response} & Time-critical triage or incident response workflow. & Must capture rapid risk stratification, escalation actions, immediate interventions, time constraints, and operational handoffs. \\
\textbf{self-study} & Structured skill acquisition or refresher training to maintain professional competence. & Must identify knowledge gaps, materials consulted, skill updates, and demonstrable evidence of retention or utility. \\
\textbf{rule-enforcement} & Application of behavioral policies or regulatory guidelines in situated contexts. & Must articulate governing policies, observed actions, context-specific exceptions, decisions, and formal appeals pathways. \\
\bottomrule
\end{tabularx}
\end{table}
\FloatBarrier

\section{Test Audit of APEX-SWE Integration Tasks}
\label{app:audit}

Table~\ref{tab:audit} presents the coding of public test suites across all 25 APEX-SWE Integration tasks according to the criteria in Section~\ref{sec:whybench}. The unit of analysis is \texttt{tests/test\_outputs.py} within each task, evaluated against the agent-facing issue in \texttt{data/issues.json}. The third column details elements within the primary deliverable that are determined by the source but lack assertions checking them against that source; the fourth column lists unchecked extraneous behaviors (such as printed confirmation lines) that do not impact coding levels. Under the public evaluation harness, tests skipped due to service errors are treated as non-failures, and tasks are marked as passed provided pytest reports no failing assertions. The supplementary CSV provides evaluated assertions and coding notes for each task, alongside records across both coding rounds and adjudication details for the 4 first-round discrepancies.

\paragraph{Definitions after the first round.}
First-round discrepancies occurred in \texttt{1-\allowbreak aws-\allowbreak s3-\allowbreak snapshots}, \texttt{5-\allowbreak localstack-\allowbreak s3-\allowbreak notifications}, \texttt{82-\allowbreak localstack-\allowbreak iam-\allowbreak user-\allowbreak creation}, and \texttt{200-\allowbreak medusa-\allowbreak list-\allowbreak products}. Three arose from ambiguities in the Section~\ref{sec:whybench} definitions, which we clarified prior to the second round. Values specified in the issue text that also appear in provided source files (such as CSV column headers) are classified as source-determined; comparing these against expected string constants counts as validating against the source. Dynamically generated runtime values (such as newly minted access key IDs) are not considered source-determined. Where the source prescribes a specific value (such as event names in received messages), verifying merely that the value belongs to an allowed set does not constitute validation against the source. The fourth discrepancy concerned field-level bindings within a single record (e.g., pairing a product ID with its corresponding title); existing definitions already classified these pairings as source-determined. For four-field coding across the three primary cases, round two populated the tested setting and evaluated result fields directly from public task assets, enumerating each public evaluation instrument and separating benchmark-provided inputs from external resources merely referenced in prompts.

{\scriptsize
\begin{longtable}{@{}>{\raggedright\arraybackslash}p{0.25\linewidth}>{\raggedright\arraybackslash}p{0.08\linewidth}>{\raggedright\arraybackslash}p{0.33\linewidth}>{\raggedright\arraybackslash}p{0.26\linewidth}@{}}
\caption{Coding of public Integration tests, ordered by evaluation tier.}\label{tab:audit}\\
\toprule
\textbf{Task} & \textbf{Level} & \textbf{Unchecked source-determined content} & \textbf{Other unchecked behavior} \\
\midrule
\endfirsthead
\toprule
\textbf{Task} & \textbf{Level} & \textbf{Unchecked source-determined content} & \textbf{Other unchecked behavior} \\
\midrule
\endhead
\texttt{278-\allowbreak plane-\allowbreak sprint-\allowbreak capacity-\allowbreak dynamodb} & \emph{none} & Assignees, capacity, hours\_allocated, utilization\_pct (none compared with Plane or capacity.yml) & --- \\
\texttt{283-\allowbreak zammad-\allowbreak sla-\allowbreak archive} & \emph{none} & Ticket rows, resolution minutes, pass flags, hit rate (none compared with Zammad or sla.json) & --- \\
\texttt{294-\allowbreak stepfunctions-\allowbreak renewal-\allowbreak orchestrator} & \emph{none} & Risk level versus opportunity probability, amount, stage; Plane issue content & --- \\
\texttt{1-\allowbreak aws-\allowbreak s3-\allowbreak snapshots} & \emph{partial} & Every uploaded row value; no assertion compares rows with users.csv & --- \\
\texttt{5-\allowbreak localstack-\allowbreak s3-\allowbreak notifications} & \emph{partial} & EventName in /app/sqs\_output.txt versus the eventName of the received message (only membership in the allowed set is checked) & --- \\
\texttt{85-\allowbreak python-\allowbreak dependency-\allowbreak checker} & \emph{partial} & Package and Current columns (never compared with requirements.txt); Latest for all rows after the first & --- \\
\texttt{200-\allowbreak medusa-\allowbreak list-\allowbreak products} & \emph{partial} & Pairing of id and title on the same line & --- \\
\texttt{222-\allowbreak espocrm-\allowbreak duplicate-\allowbreak detection} & \emph{partial} & Exact duplicate count; complete list of duplicate emails in the report & --- \\
\texttt{232-\allowbreak plane-\allowbreak issue-\allowbreak time-\allowbreak tracking} & \emph{partial} & Per-issue id, title, created\_at, completed\_at & --- \\
\texttt{234-\allowbreak medusa-\allowbreak abandoned-\allowbreak cart} & \emph{partial} & Recipient identity versus abandoned-cart emails; cart total amount (only '\$' is required in the body) & --- \\
\texttt{247-\allowbreak customer-\allowbreak journey-\allowbreak analytics} & \emph{partial} & Per-customer order\_count, support\_ticket\_count, crm\_interactions (never compared with sources); hence health\_score and classification values & --- \\
\texttt{281-\allowbreak medusa-\allowbreak zammad-\allowbreak warranty-\allowbreak validator} & \emph{partial} & Printed counts versus tickets processed; Mattermost summary content (keywords only) & --- \\
\texttt{304-\allowbreak zammad-\allowbreak bug-\allowbreak to-\allowbreak plane} & \emph{partial} & Titles of the remaining tickets; issue descriptions (first article body); correspondence between labeled issues and tickets & --- \\
\texttt{7-\allowbreak localstack-\allowbreak order-\allowbreak intake} & \emph{full} & --- & --- \\
\texttt{13-\allowbreak zammad-\allowbreak webhooks} & \emph{full} & --- & --- \\
\texttt{16-\allowbreak secret-\allowbreak rotation} & \emph{full} & --- & --- \\
\texttt{19-\allowbreak plane-\allowbreak zammad-\allowbreak tickets} & \emph{full} & --- & Printed ticket ID not checked \\
\texttt{22-\allowbreak customer-\allowbreak sentiment-\allowbreak sync} & \emph{full} & --- & --- \\
\texttt{25-\allowbreak customer-\allowbreak offboarding} & \emph{full} & --- & Printed lead id not compared with record id; changes to other leads are not rejected \\
\texttt{28-\allowbreak cross-\allowbreak app-\allowbreak sync} & \emph{full} & --- & Created/updated label not checked against prior state \\
\texttt{40-\allowbreak email-\allowbreak ticket} & \emph{full} & --- & Stdout line not checked \\
\texttt{64-\allowbreak crm-\allowbreak mailhog-\allowbreak campaigns} & \emph{full} & --- & --- \\
\texttt{76-\allowbreak mailhog-\allowbreak email-\allowbreak subject-\allowbreak fetch} & \emph{full} & --- & --- \\
\texttt{82-\allowbreak localstack-\allowbreak iam-\allowbreak user-\allowbreak creation} & \emph{full} & --- & Access key ID in the file is not compared with the key created in IAM (both are outputs of the same script) \\
\texttt{108-\allowbreak postgres-\allowbreak to-\allowbreak espocrm} & \emph{full} & --- & Printed synced count not checked; contacts for older users are not rejected \\
\bottomrule
\end{longtable}
}

\end{document}